\documentclass[]{bytedance_seed}

\usepackage[toc,page,header]{appendix}

\usepackage{tikz}
\usetikzlibrary{positioning,calc,arrows.meta,fit,backgrounds}

\usepackage{natbib}
\usepackage{amsmath, amssymb, amsthm}
\usepackage{booktabs}
\usepackage{graphicx}
\graphicspath{{figures/}{figs/}{./figs/}}
\usepackage{hyperref}
\usepackage{url}
\usepackage{enumitem}
\usepackage{listings}
\usepackage{array}
\usepackage{multirow}
\usepackage{xcolor}
\usepackage{xspace}
\usepackage{algorithm}
\usepackage{algpseudocode}
\usepackage{subcaption}
\usepackage{tcolorbox}
\usepackage{cleveref}
\usepackage{balance}
\usepackage{pifont}

\definecolor{codegreen}{rgb}{0,0.6,0}
\definecolor{codegray}{rgb}{0.5,0.5,0.5}
\definecolor{codepurple}{rgb}{0.58,0,0.82}
\definecolor{backcolour}{rgb}{0.97,0.97,0.97}

\lstdefinestyle{pythonstyle}{
    backgroundcolor=\color{backcolour},
    commentstyle=\color{codegreen},
    keywordstyle=\color{codepurple},
    numberstyle=\tiny\color{codegray},
    stringstyle=\color{codegreen},
    basicstyle=\ttfamily\small,
    breakatwhitespace=false,
    breaklines=true,
    captionpos=b,
    keepspaces=true,
    numbers=left,
    numbersep=5pt,
    showspaces=false,
    showstringspaces=false,
    showtabs=false,
    tabsize=2,
    frame=single,
    framerule=0.5pt,
    rulecolor=\color{codegray},
}
\newcommand{\vera}{\textsc{VeRA}\xspace}

\hypersetup{
    colorlinks=true,
    linkcolor=blue!70!black,
    citecolor=green!50!black,
    urlcolor=blue!70!black,
}

\usepackage{array,tabularx,float,longtable}
\newcolumntype{Y}{>{\raggedright\arraybackslash}X}
\usepackage{needspace}
\newcommand{\ve}{\mbox{\textsc{VeRA-E}}}
\newcommand{\vh}{\mbox{\textsc{VeRA-H}}}
\newcommand{\vp}{\mbox{\textsc{VeRA-H}}~\textsc{Pro}}
\title{VeRA: Verified Reasoning Data Augmentation at Scale \\
\normalsize{Programmatic Verification for Renewable Evaluation of Frontier Reasoning Models}}

\author[1,2,*, \dagger]{Zerui Cheng}
\author[1, \dagger]{Jiashuo Liu}
\author[1]{Chunjie Wu}
\author[3,2]{Jiayang Sun}
\author[2]{Jianzhu Yao}
\author[2]{Pramod Viswanath}
\author[1]{Ge Zhang}
\author[1]{Wenhao Huang}

\affiliation[1]{ByteDance Seed}
\affiliation[2]{Princeton University}
\affiliation[3]{University of Hong Kong}

\contribution[*]{Work done at ByteDance Seed}
\contribution[\dagger]{Corresponding authors}

\abstract{
Static benchmarks wear out: their problems circulate widely and can enter training data, and their scores saturate as models improve. Renewing a benchmark requires new problems with reliable answers, which are expensive to write by hand at competition level. We introduce \vera{} (\textbf{Ve}rified \textbf{R}easoning Data \textbf{A}ugmentation), in which a language model automatically compiles each benchmark problem into an \emph{executable specification}: a question template, a generator of valid inputs, and a program that computes the answer. Once its programs pass automatic checks, a specification yields many fresh, code-labeled problems at almost no cost, and people are needed only to audit. \ve{} (equivalent) draws new instances of the same computation. \vh{} (hardened) modifies the computation to create harder problems where scores saturate; because code computes every answer, the language model that writes them never has to solve them. \vp{} keeps the hardest of up to five such candidates, as ranked by a judge. Across 16 frontier models, fresh instances of AIME 2024 problems lower accuracy from 84.46\% to 70.25\%, and every model drops. GSM8K and the less-circulated Beyond-AIME, compiled by the same pipeline, stay essentially unchanged, so the drop reflects familiarity with specific problems, not an artifact of rewriting. On the near-saturated AIME 2024 II problems, \vp{} lowers accuracy from 84.91\% to 58.57\%. Expert audits back the labels, with all 220 audited fresh instances accepted and 95.1\% of hardened candidates usable after repair. Compiling all 1,319 GSM8K problems cost about USD 100. \vera{} thus makes benchmarks renewable by construction: wherever code can compute the answer, a benchmark can be refreshed whenever its problems leak or models catch up.
}

\date{\today}

\correspondence{\begin{tabular}[t]{@{}l@{\hspace{1em}}l@{}}Zerui Cheng & \email{zerui.cheng@princeton.edu}\\ Jiashuo Liu & \email{liujiashuo.77@bytedance.com}\end{tabular}}

\begin{document}
\maketitle

\begin{figure}[!t]
\centering
\includegraphics[width=\linewidth]{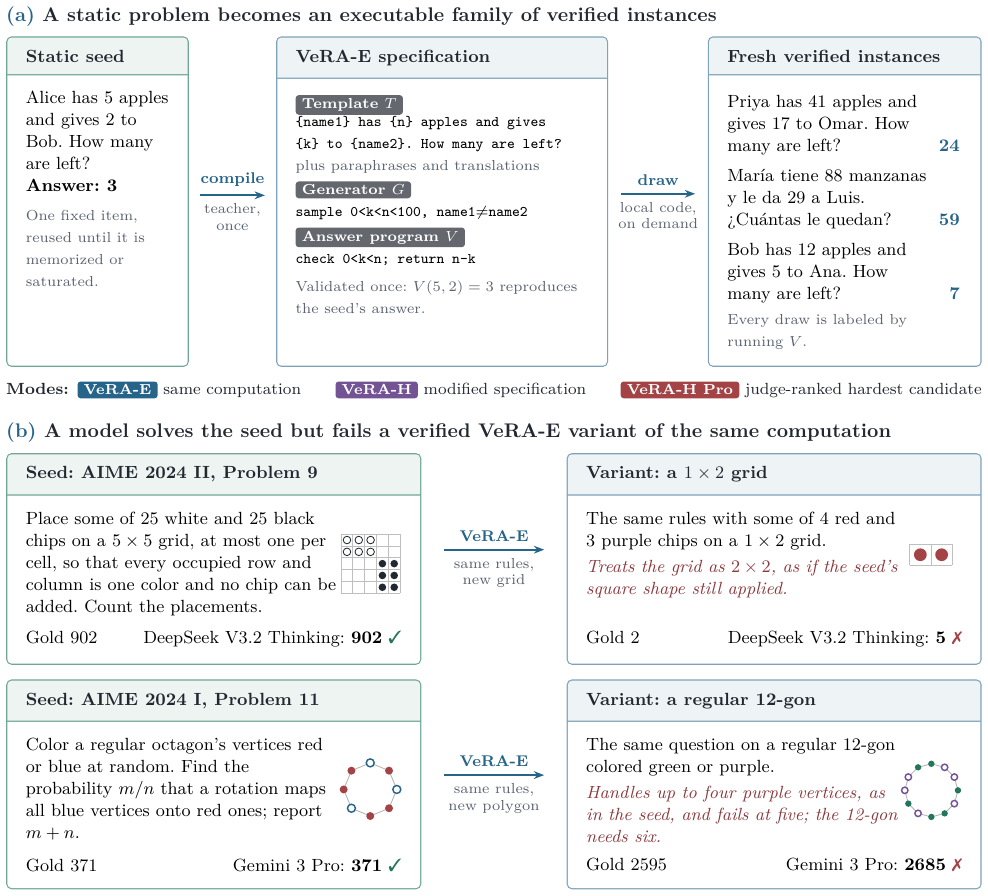}
\caption{\textbf{\vera{} at a glance.} (a) A language model, the teacher, compiles a static seed once into a \ve{} specification, which reproduces the seed's answer; local code then draws fresh instances on demand, and the answer program labels every one. \ve{} keeps the computation and varies values, names and wording; \vh{} writes a modified specification with its own generator and answer program; \vp{} keeps the judge-ranked hardest of up to five candidates. (b) Two examples behind the AIME 2024 drop from 84.46\% to 70.25\%, in which all 16 models score lower (Section~\ref{sec:fresh}): each model solves the seed but fails a verified \ve{} variant, carrying a feature of the seed's solution to an input where it no longer applies.}
\label{fig:teaser}
\end{figure}

\section{Introduction}
Static reasoning benchmarks age in two ways. Their questions circulate and can enter training data, so a high score may reflect familiarity with particular items rather than the ability to solve problems of that kind \citep{shi2023detecting,golchin2024time,sainz2023contamination}. Their difficulty also stays fixed while models improve, so the strongest models eventually cluster near the ceiling and the benchmark stops separating them \citep{jain2024livecodebench,white2024livebench}. Both problems call for new questions, and every new question needs a correct answer. Reliable labels are the central bottleneck of benchmark renewal: writing and checking competition-level problems by hand is expensive, so renewal usually means building an entirely new benchmark rather than refreshing the ones the community already relies on.

Parameterized benchmarks address part of this problem. GSM-Symbolic, Functional MATH and RV-Bench turn existing problems into templates or functions whose inputs can be resampled, and MATH-Perturb pairs problems with perturbations that change their solution method \citep{mirzadeh2024gsm,srivastava2024functional,hong2025rvbench,huang2025mathperturb}. In all of these, people write the template, function or perturbation for every problem, so coverage ends where the annotation budget ends. Automatic methods such as TemplateGSM and DARG remove the annotator, but they target grade-school mathematics or simple reasoning tasks, and none turns the problems of a competition-level benchmark into reusable families \citep{zhang2025templategsm,zhang2024darg}.

We propose a different view of what a benchmark is:
\begin{quote}
\emph{A benchmark should be a set of executable specifications that generate families of verified instances, not a static list of questions.}
\end{quote}
\vera{} (\textbf{Ve}rified \textbf{R}easoning Data \textbf{A}ugmentation) builds these specifications automatically for existing benchmarks. For each original problem (the \emph{seed}), a language model (the \emph{teacher}) writes an executable specification: a question template with slots for variable quantities, a generator that samples valid values for those slots, and an answer program that computes the label (Figure~\ref{fig:teaser}a). Each specification defines a \emph{family}: all questions that its template and generator can produce, labeled by the same answer program.

The representation supports two kinds of renewal, one for each way benchmarks age. \ve{} (\emph{equivalent}) varies inputs and wording while keeping the computation. Its fresh instances replace familiar items, and the change in accuracy from seeds to fresh instances, checked against control benchmarks, measures how much familiarity inflated the original scores. \vh{} (\emph{hardened}) changes constraints or solution structure to create harder tasks where scores have saturated, keeping evaluation at the boundary of what current models can solve. \vp{} adds selection: a difficulty judge ranks up to five validated candidates per seed and keeps the hardest. We write E, H and H Pro for short.

Validation happens once per specification. Every specification must execute correctly; an E specification must also reproduce the seed's known answer, and an H specification, which has no known answer, must pass a check by an answer judge. After that, local code produces and labels every new question, so every later draw inherits the validation. We call this \emph{correctness amortization}, and it is the core idea of \vera{}: \textbf{it decouples the scale of evaluation from the cost of reliable labels.} A validated specification labels each new instance for the price of a program run. Pipelines that rely on human grading or on agreement among models instead pay for reliability with every new item. They pay most at the frontier, where hard tasks are the most expensive to grade and the least reliable to label by agreement. People are needed only to audit items, and the audit measures how well amortization holds in each mode.

We make three contributions:
\begin{itemize}[leftmargin=*, topsep=2pt, itemsep=2pt]
\item \textbf{An automatic compiler from benchmark problems to executable families.} It covers all 1,319 GSM8K seeds for about USD 100 and all 60 AIME 2024 and 2025 seeds for about USD 11 each. Expert auditors accepted all 220 audited AIME and Beyond-AIME \ve{} items (209 new variants and 11 seeds restated as a fallback), and targeted repair makes 95.1\% of hardened candidates usable.
\item \textbf{A familiarity diagnostic with built-in controls.} Fresh instances leave accuracy unchanged on GSM8K, which is widely circulated but easy, and on Beyond-AIME, which is hard but much less widely used. On AIME 2024, which is both, they lower accuracy from 84.46\% to 70.25\%, and all 16 models score lower; AIME 2025, in circulation for less time, drops by half as much. The drop appears only on sources that are both hard and widely circulated, so it reflects familiarity with specific problems, not the pipeline. Figure~\ref{fig:teaser}b shows two such failures.
\item \textbf{Hardening with selection.} On the near-saturated AIME-2024-II problems, judge-selected \vp{} items lower accuracy from 84.91\% to 58.57\%, and no model exceeds 79\%. Because labels come from executing programs, the teacher never has to solve the tasks it creates. On every source, selection yields harder items than randomly drawn candidates. In total, the hardened sets contain 2,187 audited items.
\end{itemize}
Together, these results make the case for executable specifications as a general paradigm for evaluation: wherever code can compute the answer, a benchmark becomes renewable by construction.

Section~\ref{sec:families} defines specifications, the two renewal modes and selection; Section~\ref{sec:validation} describes how specifications and items are validated; and Section~\ref{sec:experiments} sets up the evaluation. Section~\ref{sec:results} reports construction, familiarity and hardening results, and Section~\ref{sec:uses} turns them into practical uses.

\section{Benchmarks as Executable Families}
\label{sec:families}
This section defines the specification interface, the two renewal modes built on it, and the selection step of \vp{}.

\subsection{The Specification Interface}
\begin{figure}[t]
\centering
\includegraphics[width=.92\linewidth]{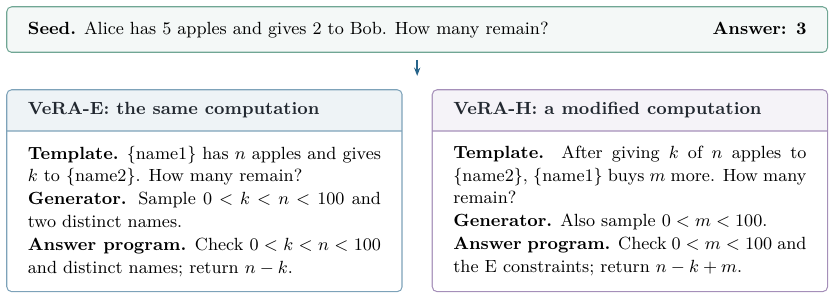}
\caption{\textbf{One seed, two kinds of executable variation.} E changes quantities and names while preserving subtraction; H adds a purchase and updates the answer program.}
\label{fig:family-example}
\end{figure}

For a seed question--answer pair $(q_i,a_i)$, the teacher proposes a specification $\mathcal S_i=(T_i,G_i,V_i)$. The template $T_i$ turns an assignment $\theta$ of values to its slots into a natural-language question; the generator $G_i$ samples such assignments using a random source $r$; and the answer program $V_i$ checks the assignment and returns a validity flag $b$ and a canonical answer $y$. An instance is drawn as follows and kept only when $b$ is true:
\begin{equation}
\theta\leftarrow G_i(r),\qquad x=T_i(\theta),\qquad (b,y)=V_i(\theta).
\end{equation}
The specification also stores slot types, constraints and, for modified tasks, the rationale for the modification. The random source $r$ is fixed by hashing the seed identifier, the generator identifier and the sampling index, so every draw is reproducible. Appendix~\ref{app:specification-details} lists all fields of a specification. Figure~\ref{fig:family-example} shows both kinds of change on a simple seed: changing quantities and names keeps the subtraction, whereas adding a purchase changes the calculation and therefore needs a new answer program. Because code labels every draw, the same specifications can also generate labeled training data (Appendix~\ref{app:exposure}).

\subsection{Family Variation and Task Modification}
\paragraph{\ve{}: variation within a computational family.}
The teacher turns the seed's computation into a parameterized family whose members, the \emph{variants}, differ from the seed in numerical values, entity names, narrative form or language. Given the seed's own values $\theta_i^{\rm seed}$, the specification must reproduce the seed answer, $V_i(\theta_i^{\rm seed})=a_i$, which anchors it to the original problem. Comparing accuracy on the seed with accuracy on fresh draws then tests whether a model applies the same reasoning to new values, cases and wordings.

\paragraph{\vh{}: modification.}
The teacher introduces new slots, constraints, dependencies or solution structure, and each modified specification carries its own generator and answer program. The teacher draws on four kinds of modification: \emph{constraint tightening} changes the admissible assignments, \emph{composition} adds intermediate quantities, \emph{parameter-regime search} moves where the generator samples, and \emph{structural generalization} replaces a special condition, such as perpendicularity, with a parameterized one, such as an angle. Each validated question that a modified specification produces is a \emph{candidate}. The pipeline generates up to five candidates per seed, and together they form the H pool.

\paragraph{\vp{}: selection.}
For H Pro, a difficulty judge ranks a separately generated pool of up to five candidates for the same seed and keeps the hardest. Because both pools come from the same process, the H pool shows what a randomly drawn candidate is like, and H Pro shows what the judge's choice adds. Figure~\ref{fig:mode-comparison} in Appendix~\ref{app:workflows} shows \ve{}, \vh{} and \vp{} on an AIME 2024 geometry problem, and the same appendix gives the construction workflows in detail.

\section{Validation and Audit}
\label{sec:validation}
Validation must establish two things: that the generated program runs on its declared inputs, and that its answer matches the question as rendered. Execution checks establish the first. The second rests on seed anchoring for E, on a judge-based answer check for H that we call \emph{noise discrimination}, and on an expert audit. Table~\ref{tab:validation} lists the layers and the evidence each supplies.

\begin{table}[t]
\centering\small
\caption{\textbf{Validation layers.} The first five layers run automatically (seed anchoring only for E, noise discrimination only for H); the expert audit checks rendered items. Each layer supplies a different kind of evidence that a label is correct.}
\label{tab:validation}
\begin{tabularx}{\linewidth}{lYY}
\toprule
Layer & Check & Evidence supplied\\
\midrule
Schema and execution & Parse fields, compile code, run sampled assignments & The specification executes through its declared interface\\
Generator coherence & Check valid yield and constraints & Sampled inputs meet the implemented requirements\\
Rendering check & Read each rendered question with a language model & No syntactic errors, contradictions or ambiguity\\
Seed anchoring (E) & Match the known seed answer & Agreement at the original input\\
Noise discrimination (H) & Judge correct and perturbed answers & Sensitivity to sampled answer errors\\
Expert item audit & Independently solve rendered problems & Semantic and label agreement on audited items\\
\bottomrule
\end{tabularx}
\end{table}

\paragraph{Automatic checks.}
Every specification must pass the same basic gates: its schema must parse, its code must compile and run on sampled inputs, and its generator must yield valid, non-degenerate instances. After rendering, a language model checks each question for syntactic errors, contradictions and ambiguity. For \ve{}, the specification must also reproduce the seed's known answer. When a check fails, the teacher receives structured feedback and revises the specification, within a budget of 20 proposals per seed. All generated code runs in child processes with time, CPU and memory limits.

\paragraph{Noise discrimination.}
Modified tasks have no known answer to check against, so an answer judge tests the label instead. For a sampled question, the judge is shown five trials in shuffled order: two present the program's answer and three a perturbed one. It judges the answer in each trial correct or incorrect, and the candidate passes only if at least four of its five verdicts are right. A judge that reliably tells the program's answer from nearby wrong answers gives evidence, separate from execution, that the program labels the rendered question correctly. On 50 calibration items, this threshold gave one false acceptance and three false rejections, whereas requiring all five gave zero and fourteen (Appendix~\ref{app:noise}).

\paragraph{Expert audit.}
Six STEM PhD auditors solved rendered items without seeing the generated code, compared their answers with the program's output, and assessed whether the question was clear and correctly rendered; for \ve{} items they also checked that the question preserves the seed's computation. When an auditor's answer disagreed with the program's, a senior auditor adjudicated, and unresolved cases were rejected. Because every item is tied to its specification, an audit failure can be localized to the wording, the answer program or the parameter domain, and the pipeline repairs it at its source (Section~\ref{sec:construction}).

\section{Experimental Setup}
\label{sec:experiments}
\paragraph{Models and data.}
We evaluate 16 models from nine families: GPT-5 and GPT-5.1 (high and chat), Gemini 2.5 Pro and 3 Pro, Claude Sonnet 4.5, DeepSeek V3.1 and V3.2, Seed-1.6 (three variants), Kimi K2 (Thinking and 0905), Qwen3-Max, GLM-4.6 and MiniMax M2. Fresh instances are drawn for GSM8K \citep{cobbe2021training}, AIME 2024 and 2025, and Beyond-AIME \citep{bytedance_seed_2025_beyondaime}. Hardening uses historical AIME (1983--2001), AIME-2024-II (14 problems from AIME 2024 II), Beyond-AIME and AMO-Bench, a set of olympiad problems filtered to challenge strong reasoning models \citep{an2025amo}; H Pro is built for the last three. The teacher and both judges (answer and difficulty) are GPT-5 (high reasoning effort), which is also one of the evaluated models; Section~\ref{sec:fresh} repeats the AIME comparison with two other teachers.

\paragraph{Scoring.}
The evaluated models receive zero-shot reasoning prompts with boxed-answer extraction \citep{wei2022chain} and answer each problem five times with independent sampling. Let $Y_{mjk}$ indicate whether response $k$ of model $m$ to item $j$ is correct. The score of model $m$ on an item set $\mathcal D$ is
\begin{equation}
 S_m(\mathcal D)=\frac{100}{5|\mathcal D|}\sum_{j\in\mathcal D}\sum_{k=1}^{5}Y_{mjk},\label{eq:avg5}
\end{equation}
which we call Avg@5. The panel mean averages $S_m$ equally over the 16 models, and Appendix~\ref{app:models} lists all 16 models and their scores. We report 95\% t-intervals over the 16 per-model differences (for example, \ve{} minus seed), treating models as paired units (Appendix~\ref{app:estimands}).

\paragraph{Evaluation sets.}
Fresh-instance results use every seed and its variants: two variants per seed for GSM8K and each AIME year, and one per seed for Beyond-AIME. Of the 100 Beyond-AIME variants, 11 restate their seed because no specification passed the automatic checks within the proposal budget (Table~\ref{tab:synthesis-yield}). The 220 AIME and Beyond-AIME \ve{} items were audited by experts, whereas GSM8K relies on the automatic checks alone. Hardening results come at two stages: the \emph{raw} pools contain every candidate before audit, and the \emph{Verified} sets keep the items accepted after audit and repair (Table~\ref{tab:hardening}); H and H Pro are audited separately. Hardening accuracies refer to the Verified sets unless marked raw; historical AIME has no H Pro set and was evaluated only on its raw H pool. Appendix~\ref{app:gpqa} reports a GPQA-Diamond extension beyond computable answers.

\section{Results}
\label{sec:results}
We ask three questions: whether \vera{} builds valid families at scale and low cost (Section~\ref{sec:construction}), whether fresh instances expose familiarity (Section~\ref{sec:fresh}), and whether hardening restores headroom where seeds saturate (Section~\ref{sec:hardening-results}).\footnote{Kimi K2 Thinking scores erratically across sources, for example 15.1\% on the raw AIME-2024-II H pool but 45.7\% on the harder H Pro items. Excluding it leaves every conclusion unchanged (Appendix~\ref{app:robustness}).}

\subsection{Automatic Construction at Scale}
\label{sec:construction}
No person writes a template, generator or answer program. Specification synthesis cost about USD 100 for the 1,319 GSM8K seeds and about USD 11 per seed for AIME problems, after which each further instance costs one local program run. Table~\ref{tab:synthesis-yield} shows the resulting coverage: every GSM8K, AIME 2024 and AIME 2025 seed yields a specification that passes all automatic checks within 20 proposals, as do 89 of 100 Beyond-AIME seeds.

\begin{table}[t]
\centering\small
\caption{\textbf{Construction yield.} E: seeds whose specification passes all automatic checks on the first proposal or within 20, and seeds that fall back to restating the seed. H: seeds by the number of candidates in the H pool before audit (target five); historical AIME yields 1,321 candidates from 265 seeds (Appendix~\ref{app:extended-models}).}
\label{tab:synthesis-yield}
\begin{tabular*}{\linewidth}{@{\extracolsep{\fill}}lrrrr@{}}
\toprule
E source & Seeds & First proposal & Within 20 & Fallback\\
\midrule
GSM8K & 1,319 & 1,319 & 1,319 & 0\\
AIME 2024 & 30 & 29 & 30 & 0\\
AIME 2025 & 30 & 27 & 30 & 0\\
Beyond-AIME & 100 & 63 & 89 & 11\\
\midrule
H source & Seeds & Five candidates & Four & Zero\\
\midrule
AIME-2024-II & 14 & 14 & 0 & 0\\
Beyond-AIME & 100 & 100 & 0 & 0\\
AMO-Bench & 50 & 48 & 1 & 1\\
\bottomrule
\end{tabular*}
\end{table}

Fresh instances need human attention only for auditing, and the audit finds nothing to fix: experts accepted all 220 AIME and Beyond-AIME \ve{} items. With zero errors among the 209 generated variants, the one-sided 95\% upper bound on the error rate of a new draw is 1.4\%, or 2.0\% if each of the 149 families counts once (Appendix~\ref{app:audit}). New E draws from these sources can therefore be used without review: correctness amortization holds in practice. GSM8K, which relies on the automatic checks alone, also appears to yield cleaner items. In a separate single-response probe, the share of questions that both GPT-5 and Gemini 2.5 Pro miss, a proxy for flawed statements or labels, falls from 2.12\% to 0.76\% after rewriting (Appendix~\ref{app:gsm-probe}).

Hardened candidates change the computation itself and have no seed answer to anchor them, so the audit matters most there. The initial audit accepts 1,733 of 2,299 hardened candidates (75.4\%). Targeted repair recovers 454 more, raising the usable yield to 95.1\% and the hardened sets to 2,187 items (Table~\ref{tab:hardening}). The audit catches three kinds of error that execution checks cannot: an answer program that mishandles an integer or rational boundary, a template that allows two readings although its code computes one answer, and a sampled parameter regime that passes the automatic non-degeneracy check yet trivializes the intended challenge. Repair acts on the source of each error: the pipeline fixes the code behind an arithmetic error, rewrites an ambiguous question, or tightens the domain constraints that allowed a degenerate assignment. Each corrected candidate then passes through validation and audit again. The most demanding source, AMO-Bench, needs the most correction. Correctness amortization therefore holds fully for E, whereas hardened items still need an audit before use.

\begin{table}[t]
\centering\small
\caption{\textbf{Construction quality of the hardened sets.} Candidates form the raw pool; Initial pass counts candidates accepted by the first expert audit, and Final accepted gives the Verified set after repair and re-audit. AIME 1983--2001 is historical AIME. Totals include historical AIME as well as AIME-2024-II, Beyond-AIME and AMO-Bench.}
\label{tab:hardening}
\begin{tabular*}{\linewidth}{@{\extracolsep{\fill}}llrrr@{}}
\toprule
Source & Mode & Candidates & Initial pass & Final accepted\\
\midrule
AIME 1983--2001 & H & 1,321 & 1,043 & 1,284\\
AIME-2024-II & H & 70 & 63 & 69\\
Beyond-AIME & H & 500 & 365 & 466\\
AMO-Bench & H & 244 & 155 & 217\\
AIME-2024-II & H Pro & 14 & 11 & 14\\
Beyond-AIME & H Pro & 100 & 69 & 93\\
AMO-Bench & H Pro & 50 & 27 & 44\\
\midrule
Total & & 2,299 & 1,733 & 2,187\\
\bottomrule
\end{tabular*}
\end{table}

\subsection{Fresh Instances Expose Benchmark Familiarity}
\label{sec:fresh}

\begin{table}[t]
\centering\small
\setlength{\tabcolsep}{4pt}
\caption{\textbf{Fresh instances (\ve{}) versus seeds.} Panel-mean Avg@5 accuracy (\%) over 16 models. Change is \ve{} minus seed (points); the 95\% CI is a paired t-interval over models; Lower counts models whose accuracy falls; $\rho$ is Spearman's correlation between seed and \ve{} scores. The 220 AIME and Beyond-AIME \ve{} items were audited by experts; GSM8K relies on the automatic checks alone. $^\dagger$Includes 11 fallback items that restate the seed because no specification passed the automatic checks within 20 proposals.}
\label{tab:equivalent}
\begin{tabular*}{\linewidth}{@{\extracolsep{\fill}}lrrrrrrrr@{}}
\toprule
Benchmark & Seeds & Variants & Seed & \ve{} & Change & 95\% CI & Lower & $\rho$\\
\midrule
GSM8K & 1,319 & 2,638 & 94.85 & 95.20 & +0.35 & [$-$0.9, +1.6] & 1/16 & 0.71\\
Beyond-AIME & 100 & 100$^\dagger$ & 58.34 & 57.30 & $-$1.04 & [$-$3.0, +0.9] & 10/16 & 0.87\\
AIME 2025 & 30 & 60 & 79.21 & 72.13 & $-$7.08 & [$-$9.8, $-$4.3] & 16/16 & 0.88\\
AIME 2024 & 30 & 60 & 84.46 & 70.25 & $-$14.21 & [$-$17.0, $-$11.4] & 16/16 & 0.81\\
\bottomrule
\end{tabular*}
\end{table}

Table~\ref{tab:equivalent} compares seed and \ve{} accuracy on four benchmarks that differ in how hard their problems are for current models and in how widely those problems have circulated. Every E specification reproduces its seed's answer, and every AIME and Beyond-AIME item passed the expert audit, so a change in accuracy measures how models handle the same computation on new inputs; the controls below separate familiarity from any difficulty the new inputs add. GSM8K is the most widely circulated of the four but is easy for every model in the panel. Fresh instances leave its mean unchanged (+0.35 points), and 15 of 16 models score slightly higher on the variants.

Beyond-AIME is the opposite case: its problems match or exceed the hardest AIME problems \citep{bytedance_seed_2025_beyondaime}, but the benchmark is much less widely used, so it serves as a control that keeps the difficulty but not the circulation. Fresh instances again leave its mean nearly unchanged ($-$1.04 points). The 11 fallback items repeat their seeds and could dilute a drop, but attributing the whole change to the 89 compiled families still gives only about $-$1.2 points.

\begin{figure}[t]
\centering
\includegraphics[width=\linewidth]{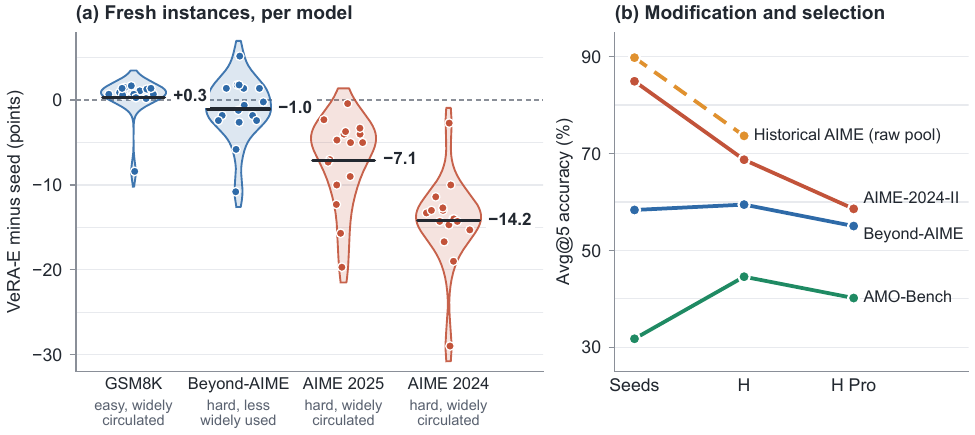}
\caption{\textbf{Fresh instances and hardened sets across the 16-model panel.} (a) Change from seed to \ve{} accuracy across the 16 models: each violin shows the smoothed distribution, each dot is one model, and the bar marks the panel mean; only the hard, widely circulated AIME sets drop. (b) Panel-mean Avg@5 accuracy on seeds, H and H Pro (Verified sets; historical AIME shows its raw H pool and has no H Pro set). Model-level values are in Appendix~\ref{app:models} for (a) and, for the raw pools, in Appendix~\ref{app:extended-models} for (b).}
\label{fig:diagnostics}
\end{figure}

AIME is both hard and widely circulated, and its two years are the only sources whose accuracy clearly falls (Table~\ref{tab:equivalent}; Figure~\ref{fig:diagnostics}a). On AIME 2024 the drop is 14.21 points (95\% CI 11.4 to 17.0) and affects every model. Measured against the Beyond-AIME control, which passed through the same pipeline, the excess drop is 13.2 points (9.9 to 16.4) and holds for all 16 models.

The two AIME years sharpen the picture. AIME 2025, which has circulated for a shorter time, drops by 7.08 points, half as much as AIME 2024. Edit size does not explain the difference, because the two years are edited by similar amounts: a perturbation score that combines numerical and textual edit distance averages 0.28 for 2024 and 0.30 for 2025 (Kolmogorov--Smirnov $D=0.15$, $p=0.51$; Figure~\ref{fig:perturbation} in Appendix~\ref{app:perturbation}). The ordering does not depend on the teacher. When Gemini 3 Pro or Claude Opus 4.6 constructs the variants instead, AIME 2024 again drops about twice as much as AIME 2025 (12.8 versus 6.9 points and 15.1 versus 6.3, respectively).

Together, these results form the signature of familiarity. The drop is confined to the hard, widely circulated sources, is twice as large for the year that has circulated longer, and persists under three different teachers. Fresh instances thus detect familiarity with specific AIME problems from model behavior alone.

Figure~\ref{fig:teaser}b shows what the drop looks like for individual problems, and Appendix~\ref{app:worked-failures} gives the full statements. In both examples the model carries over a feature of the seed's solution (a square grid, or a case split that stops at four) to an input where it no longer applies. Because the answer program fixes the label for every input, each failure can be traced to the parameter change that exposed it, and further draws from the same family can locate the unstable regime.

\subsection{Hardening Restores Headroom Where Seeds Saturate}
\label{sec:hardening-results}
Modification has its largest effect where renewal is most needed (Figure~\ref{fig:diagnostics}b; Table~\ref{tab:raw-verified}). It lowers accuracy by about 16 points on the two saturated sources, historical AIME (raw pool; Table~\ref{tab:historical-aime} in Appendix~\ref{app:extended-models}) and AIME-2024-II, whose seeds score 89.8\% and 84.9\%. It leaves Beyond-AIME almost unchanged (58.3\% to 59.5\%) and raises accuracy on AMO-Bench from 31.8\% to 44.6\%. Across these four sources, H accuracy roughly follows $30+0.47\times$ seed accuracy, a line that crosses the diagonal, where H equals the seed, at about 57\%: sources whose seeds score well above it become harder, those well below become easier, and Beyond-AIME, near the crossing, barely moves. Modified tasks thus settle near 57\% panel accuracy, the \emph{certifiable difficulty level} of the pipeline: the accuracy at which tasks whose labels code can compute and a judge can check tend to settle. This level lies well below the accuracy of the saturated AIME seeds. The same fit for H Pro over its three sources crosses at 47.5\%, so selection lowers this level by about ten points.

Three factors plausibly explain why hardening makes AMO-Bench (seed accuracy 32\%) easier rather than harder. First, its seeds were filtered to challenge strong reasoning models \citep{an2025amo}, so regenerated problems regress toward the mean. Second, the answer program must compute every label, which pushes insight-driven olympiad problems toward parameter ranges that admit enumeration. Third, noise discrimination admits only tasks whose answers the judge can reliably tell from nearby wrong ones, a bar that the hardest tasks are the most likely to miss. In our Beyond-AIME synthesis runs, the check rejected 44.0\% of candidates that had passed execution.

\begin{table}[t]
\centering\small
\setlength{\tabcolsep}{4pt}
\caption{\textbf{Hardening and selection.} Avg@5 accuracy (\%) over the 16-model panel on seeds, raw pools (every candidate, before audit) and Verified sets. The last three columns give H Pro minus H on the raw pools (points), its paired 95\% interval and the number of models that score lower. On AIME-2024-II, raw and Verified H Pro coincide because repair changed only three items, which no model solved before or after repair (Appendix~\ref{app:automatic}).}
\label{tab:raw-verified}
\begin{tabular*}{\linewidth}{@{\extracolsep{\fill}}lrrrrrrrr@{}}
\toprule
 & & \multicolumn{2}{c}{Raw pool} & \multicolumn{2}{c}{Verified} & \multicolumn{3}{c}{H Pro $-$ H, raw}\\
\cmidrule(lr){3-4}\cmidrule(lr){5-6}\cmidrule(lr){7-9}
Source & Seeds & H & H Pro & H & H Pro & Change & 95\% CI & Lower\\
\midrule
AIME-2024-II & 84.91 & 67.80 & 58.57 & 68.71 & 58.57 & $-$9.2 & [$-$15.4, $-$3.1] & 15/16\\
Beyond-AIME & 58.34 & 58.78 & 53.98 & 59.45 & 55.02 & $-$4.8 & [$-$6.3, $-$3.3] & 15/16\\
AMO-Bench & 31.75 & 42.97 & 38.38 & 44.56 & 40.16 & $-$4.6 & [$-$5.9, $-$3.3] & 15/16\\
\bottomrule
\end{tabular*}
\end{table}

Selection then adds difficulty on every source. Every AIME-2024-II and Beyond-AIME seed contributes exactly five raw H candidates, as do 48 of the 50 AMO-Bench seeds, so the H mean is in effect the expected accuracy of a randomly drawn candidate. The gap between H Pro and H therefore measures the contribution of selection (Section~\ref{sec:families}; last columns of Table~\ref{tab:raw-verified}). On the raw pools, H Pro is harder than H on all three sources, and 15 of 16 models follow this ordering on each. On the Verified sets, the mean gaps are 10.1, 4.4 and 4.4 points (Appendix~\ref{app:automatic}).

Hardening also adds difficulty beyond freshness. Part of the gap between the AIME-2024-II seeds and H Pro comes from freshness alone, because fresh instances of the same AIME 2024 computations are already harder for these models (Section~\ref{sec:fresh}). Three comparisons separate the two effects. First, the selection gap above holds freshness fixed, since H and H Pro are equally new to every model. Second, fresh instances alone bring the full AIME 2024 set, which contains AIME-2024-II, to 70.25\%, still 11.7 points above H Pro. Third, on Beyond-AIME, where freshness costs only about one point, H Pro lies 4.4 points below the seeds on the raw pool (95\% CI 1.7 to 7.0) and 3.3 points below on the Verified set.

\section{Using \vera{} in Evaluation}
\label{sec:uses}
Fresh instances and hardened sets have two direct uses, which together suggest a reporting protocol.

\paragraph{Restoring separation.}
Hardened items restore separation where seeds saturate. On AIME-2024-II seeds, seven of 16 models score at least 90\% and the top four lie within 1.4 points; on the raw H Pro items, no model exceeds 79\% and the top four span 11.5 points. On historical AIME, 13 models reach 90\% on the seeds and none on the raw H pool.

\paragraph{Revealing distorted rankings.}
Fresh instances also expose rankings that familiar items distort. Seed and E scores correlate well overall (Spearman's $\rho=0.81$ on AIME 2024), yet individual models move substantially: GLM-4.6 falls from a tie for fourth to twelfth; MiniMax M2 loses 29.0 points, twice the panel mean, and also has the largest drop on AIME 2025. A robustness test at grade-school level would miss this entirely: on GSM8K, fresh instances lower the score of only one of the 16 models.

\paragraph{A reporting protocol.}
For any benchmark compiled into executable specifications, we recommend the following protocol:
\begin{enumerate}[leftmargin=*, topsep=2pt, itemsep=1pt]
\item report seed accuracy together with E accuracy;
\item add a robustness statistic, such as the seed-to-E change, the rate at which a model that solves a seed also solves its variants, or the seed--E rank correlation;
\item include controls, such as GSM8K and Beyond-AIME, that separate familiarity from pipeline effects;
\item where seeds saturate, report Verified H Pro accuracy alongside seed accuracy to restore separation;
\item when task families also serve as training data, document the exposure, and hold out either new draws from seen families, to test new instances of known computations, or entire families, to test new computations (Appendix~\ref{app:exposure}).
\end{enumerate}

\section{Related Work}
\label{sec:related}
\begin{table}[t]
\centering\footnotesize
\setlength{\tabcolsep}{2pt}
\caption{\textbf{Construction of related benchmarks.} Coverage refers to the source problems turned into families or perturbations. Redraw: whether new instances can be drawn on demand (Functional MATH releases periodic snapshots). Hardening: how harder variants are produced.}
\label{tab:related}
\begin{tabular*}{\linewidth}{@{\extracolsep{\fill}}llllll@{}}
\toprule
Approach & Written by & Coverage & Hardest source & Redraw & Hardening\\
\midrule
GSM-Symbolic & Humans & 100 templates & GSM8K & Yes & Added clauses\\
Functional MATH & Humans & 41.2\% of MATH & MATH & Snapshots & None\\
RV-Bench & Humans & 230 of 500 sampled & MATH/LeetCode & Yes & None\\
MATH-Perturb & Humans & 279 pairs & MATH level 5 & No & Manual\\
TemplateGSM & Automatic & New templates & Grade school & Yes & None\\
DARG & Automatic & Per item & GSM8K, BBH & No & Graph depth\\
\vera{} (ours) & Automatic & 100\% GSM8K, AIME 2024/25; & AMO-Bench & Yes & Modification\\
 & & 89\% Beyond-AIME & & & + selection\\
\bottomrule
\end{tabular*}
\end{table}

\paragraph{Parameterized and perturbed benchmarks.} GSM-Symbolic, Functional MATH, RV-Bench, Putnam-AXIOM and VAR-MATH parameterize existing problems \citep{mirzadeh2024gsm,srivastava2024functional,hong2025rvbench,gulati2025putnam,yao2025varmath}, while GSM-Plus, GSM-Ranges and MATH-Perturb perturb statements, numerical ranges or solution methods \citep{li2024gsmplus,shrestha2025gsmranges,huang2025mathperturb}. They share \vera{}'s view of a problem as a family, but each template, function or perturbation is written by hand, which bounds their coverage (Table~\ref{tab:related}).

\paragraph{Automatic generation.} TemplateGSM asks GPT-4 for parameterized meta-templates with code solutions, and DyVal, DARG and AutoLogi generate graph- or logic-based tasks with programmatic checks \citep{zhang2025templategsm,zhu2024dyval,zhang2024darg,liu2025autologi}. These methods synthesize new tasks or perturb grade-school and simple reasoning items, and none compiles an existing competition-level benchmark; \vera{} does, anchors each family to its seed, and adds hardening with selection.

\paragraph{Freshness and familiarity.} Exposure analyses and live benchmarks motivate fresh test material \citep{shi2023detecting,golchin2024time,sainz2023contamination,jain2024livecodebench,white2024livebench}. PaCoST compares a model's confidence on original and rephrased items \citep{zhang2024pacost}, and GSM1k measures performance on newly written grade-school problems against GSM8K \citep{zhang2024gsm1k}. \ve{} measures familiarity at competition level without hand-written problems: its answers are computed by programs, and controls such as Beyond-AIME separate familiarity from difficulty.

\paragraph{Execution and verification.} PAL and Program of Thoughts execute programs to solve problems \citep{gao2023pal,chen2023program}, and EvalPlus generates further tests for code \citep{liu2023evalplus}. In \vera{}, execution instead labels new benchmark instances, and program checks, audits and repair then validate them. Learned verifiers and multi-agent debate supply complementary checks \citep{lightman2023verify,du2023debate}. In a side study, structured debate raises the precision of answer acceptance from 55.1\% to 92.5\% (Appendix~\ref{app:debate}), so debate could replace the single answer judge. Appendix~\ref{app:research-context} discusses further related work.

\section{Limitations}
\vera{}'s labels are executable only when the answer is a computable function of structured inputs. This covers competition mathematics with numerical answers and extends naturally to code reasoning, constraint puzzles, quantitative science and planning with a plan validator. Conceptual multiple-choice, commonsense and proof-based tasks need another source of labels, as in the GPQA-Diamond extension of Appendix~\ref{app:gpqa}. Hardening settles near the pipeline's certifiable difficulty level (Section~\ref{sec:hardening-results}), so it suits saturated benchmarks best, and a stronger teacher or judge may move that level toward harder tasks. The study fixes the candidate budget at five and uses GPT-5, itself an evaluated model, as the only teacher for H and H Pro and as the only judge; varying these choices is a natural extension.

Finally, fresh instances detect familiarity from model behavior, whereas confirming contamination would require the training data, which only model developers hold. Compiling a problem set published after every panel model's training cutoff, alongside an older one, would give a confirmatory test, since the diagnostic predicts a drop only on the older set. Because compilation is automatic, this test is straightforward to run on future panels.

\section{Conclusion}
Static reasoning benchmarks lose value as their problems become familiar and their scores saturate. \vera{} offers a different view: a benchmark is a set of executable specifications that generate verified instances on demand, which makes renewal automatic and auditable. A language model compiles each problem into a family, code labels every new draw, and an expert audit certifies the competition-level items. \ve{} exposes familiarity that static scores hide, \vh{} with \vp{} selection restores headroom where seeds saturate, and the same specifications can also generate training data with executable labels (Appendix~\ref{app:exposure}). By making renewal a property of the benchmark itself, \vera{} provides infrastructure for evaluation that stays fresh, auditable and scalable as models advance.
\label{sec:main-end}

\bibliographystyle{plainnat}
\bibliography{references}
\clearpage
\beginappendix
The appendices are grouped by topic. Appendices~\ref{app:specification-details}--\ref{app:audit} describe how specifications are built and validated: the specification and construction loop (\ref{app:specification-details}), worked examples on an AIME geometry seed (\ref{app:workflows}), and noise discrimination and the expert audit (\ref{app:audit}). Appendix~\ref{app:estimands} defines the scores, intervals and fitted quantities. Appendices~\ref{app:models}--\ref{app:robustness} give model-level results and further analyses for fresh instances and hardening. Appendices~\ref{app:gpqa}--\ref{app:exposure} cover extensions: a GPQA-Diamond extension, structured debate as an answer judge, and training use. Appendix~\ref{app:research-context} discusses related work in more depth.

\section{Specification and Construction Details}
\label{app:specification-details}
This appendix describes what a specification stores, how the teacher proposes and repairs it, and how many seeds yield one within the proposal budget.

\subsection{A Complete Specification}
A specification stores everything needed to regenerate, label and audit its family (Table~\ref{tab:fields}): the slot names and types, the admissible assignments, a rendering template, a generator and an answer program. An E specification also stores the base assignment that links it to its seed, and an H specification records which condition changed and why. These fields connect the question a model sees with the computation that produces its answer, make every draw reproducible, and let every audit failure be traced to its source.
\begin{table}[htbp]
\centering\small
\caption{\textbf{Specification fields.} Every accepted specification stores the fields that apply to its mode.}
\label{tab:fields}
\begin{tabularx}{\linewidth}{lY}
\toprule
Field & Role\\
\midrule
Identifiers & Seed identifier, generator identifier, version and sampling index, which fix each draw's random source.\\
Slot schema & Defines quantities, types, ranges, and relationships among inputs.\\
Template & Renders the assignment into the question presented to the evaluated model.\\
Generator & Samples assignments using the supplied random-number source.\\
Answer program & Checks an assignment and computes its canonical answer.\\
Base assignment & Recreates the seed case for E anchoring.\\
Modification rationale & Identifies the intended change in constraints or computation for H.\\
Validation record & Associates the specification and sampled items with execution, judgment, and audit outcomes.\\
\bottomrule
\end{tabularx}
\end{table}

A small example shows what the interface requires. Consider a family that asks for the total in $a$ rows of $b$ objects plus $c$ additional objects, where $a$ and $b$ are integers from 2 through 20, $c$ ranges from 1 through 10, and $a\ne b$. A coherent generator draws $a$, selects $b$ from the remaining permitted values, and draws $c$; the answer program checks these constraints and returns $ab+c$, so the assignment $(5,4,3)$ renders a problem with answer 23. Every constraint must appear both in the generator and in the validity check, and the template must state every quantity that the answer program uses.

\subsection{Proposal, Execution and Repair}
The teacher proposes a structured specification rather than a single free-form problem. The pipeline parses its schema, compiles its functions and runs them on sampled assignments. An E specification must also reproduce the seed answer at its base assignment; an H specification has no known answer and must instead pass noise discrimination (Appendix~\ref{app:noise}). After rendering, a language model checks each question for syntactic errors, contradictions and ambiguity (Figures~\ref{fig:workflow-e} and~\ref{fig:workflow-h}). When a check fails, the teacher receives concrete feedback, such as a compilation error, an invalid assignment, a generator that yields no valid instance, or an answer mismatch, and revises the specification.

Synthesis allows up to 20 proposals per seed, tests five sampled assignments per proposal, and aims for five accepted variants (E) or candidates (H) per seed. Generated code runs in child processes with timeouts and CPU and memory limits, and generators receive their randomness through a seeded interface rather than from global state, which keeps every draw reproducible.

When no proposal within the budget passes, an explicit fallback restates the seed, and the item's metadata marks this path. A fallback item repeats the seed's question and answer, so it is not a new instance; flagging it keeps coverage separate from variation.

\subsection{Synthesis Yield}
Repeated proposals recover most seeds that fail on the first try (Table~\ref{tab:synthesis-yield}): both AIME years reach full coverage, and Beyond-AIME rises from 63 to 89 seeds, with fallback covering the remaining 11. For H, every AIME-2024-II and Beyond-AIME seed yields all five candidates, and the AMO-Bench counts give $48\times5+1\times4=244$ candidates.

\FloatBarrier
\section{Construction Workflows}
\label{app:workflows}
\begin{figure}[!ht]
\centering
\includegraphics[width=.85\linewidth]{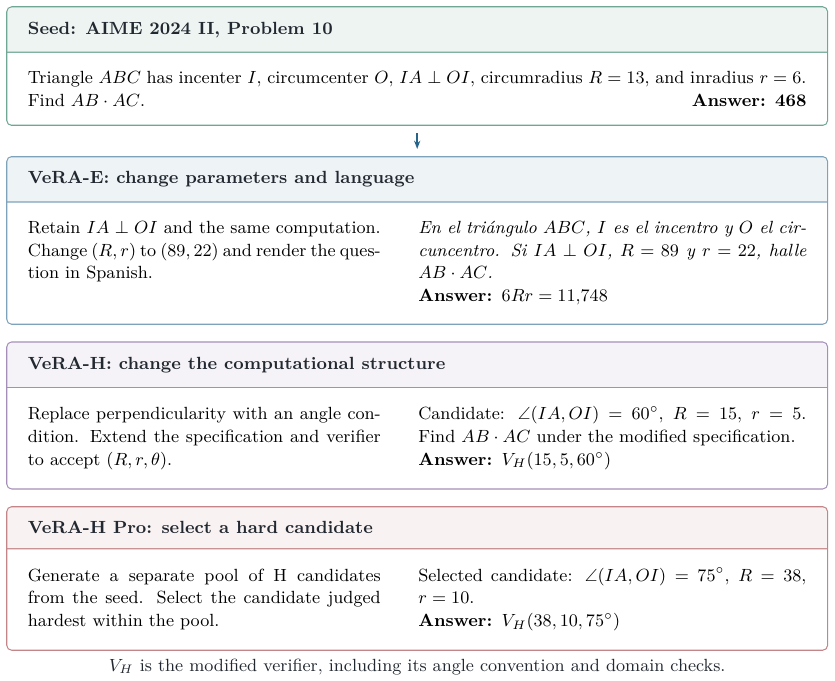}
\caption{\textbf{\ve{}, \vh{} and \vp{} on one geometry seed (AIME 2024 II, Problem 10).} E changes parameters and language, H replaces the perpendicularity condition with an angle, and H Pro selects the hardest candidate from a separately generated pool. The E example keeps the perpendicularity condition and has answer $6Rr=11{,}748$.}
\label{fig:mode-comparison}
\end{figure}

\begin{figure}[p]
\centering
\includegraphics[width=\linewidth,height=.84\textheight,keepaspectratio]{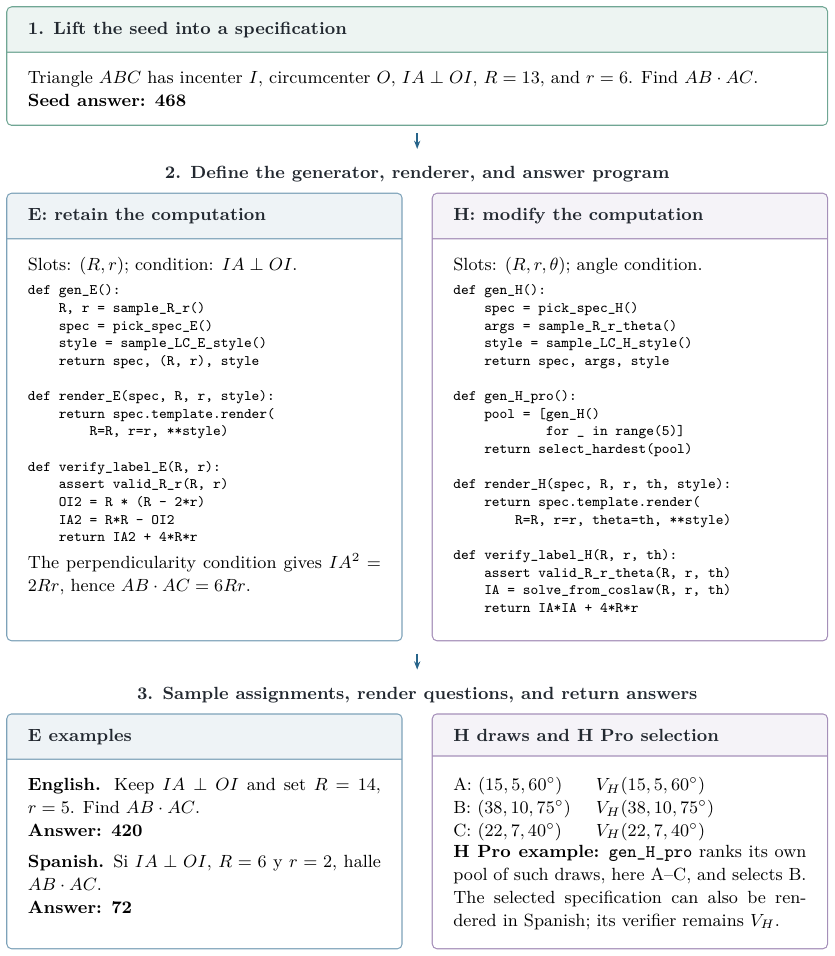}
\caption{\textbf{From a geometry seed to executable families.} The pipeline has three stages: lifting the seed into a specification; defining the generator, renderer and answer program; and sampling assignments to render questions and return answers. The E family keeps the perpendicularity condition, whereas the H family turns it into an angle parameter. The code blocks show how sampling and answer computation are separated. The H branches are modified specifications, which still pass through validation and expert audit before they enter the Verified set. For brevity, the H Pro example ranks draws A--C; in the pipeline, H Pro ranks its own, separately generated pool.}
\label{fig:synthesis-pipeline}
\end{figure}

\begin{figure}[p]
\centering
\includegraphics[width=\linewidth,height=.84\textheight,keepaspectratio]{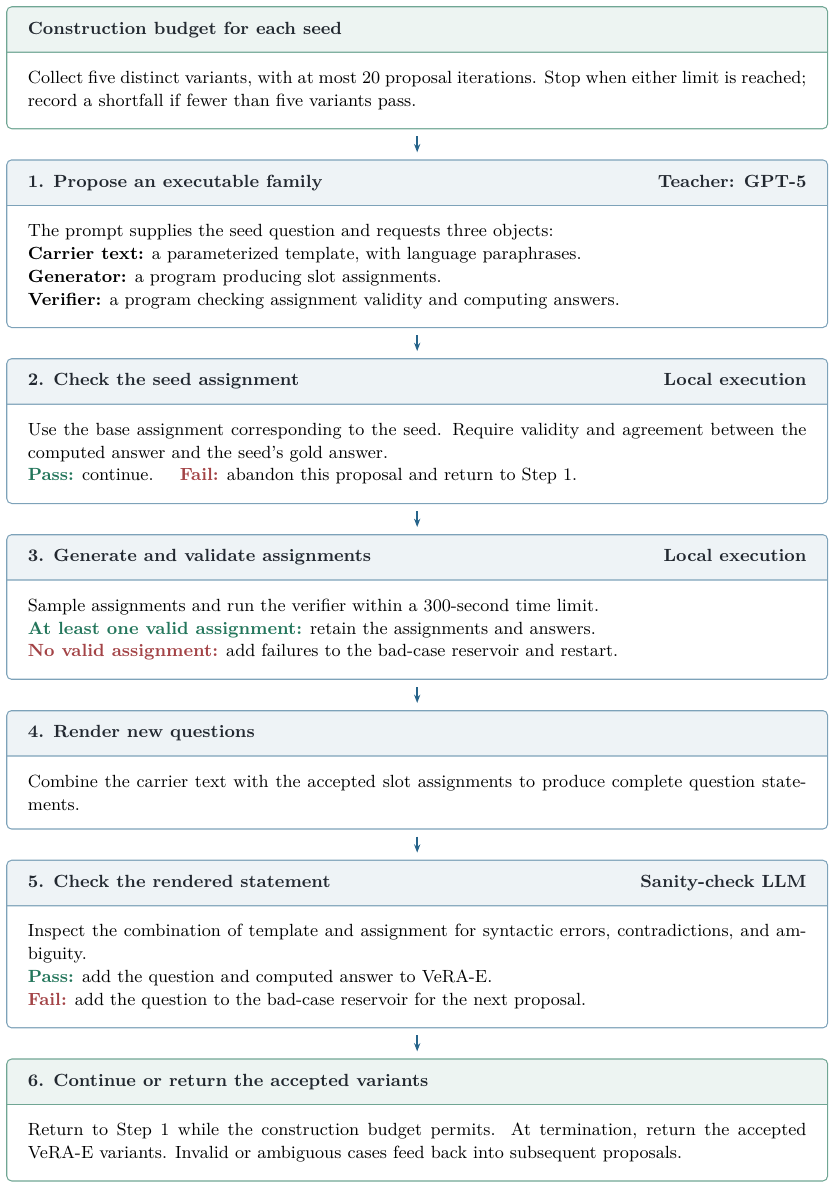}
\caption{\textbf{The E construction loop in detail.} Seed anchoring, generation of valid assignments and rendering checks all return feedback to the teacher. The loop stops after five accepted variants or when the 20-proposal budget is exhausted; two variants per seed are evaluated for GSM8K and each AIME year, and one for Beyond-AIME. ``Carrier text'' is the template, ``verifier'' the answer program, and the sanity-check LLM performs the rendering check.}
\label{fig:workflow-e}
\end{figure}

\begin{figure}[p]
\centering
\includegraphics[width=\linewidth,height=.84\textheight,keepaspectratio]{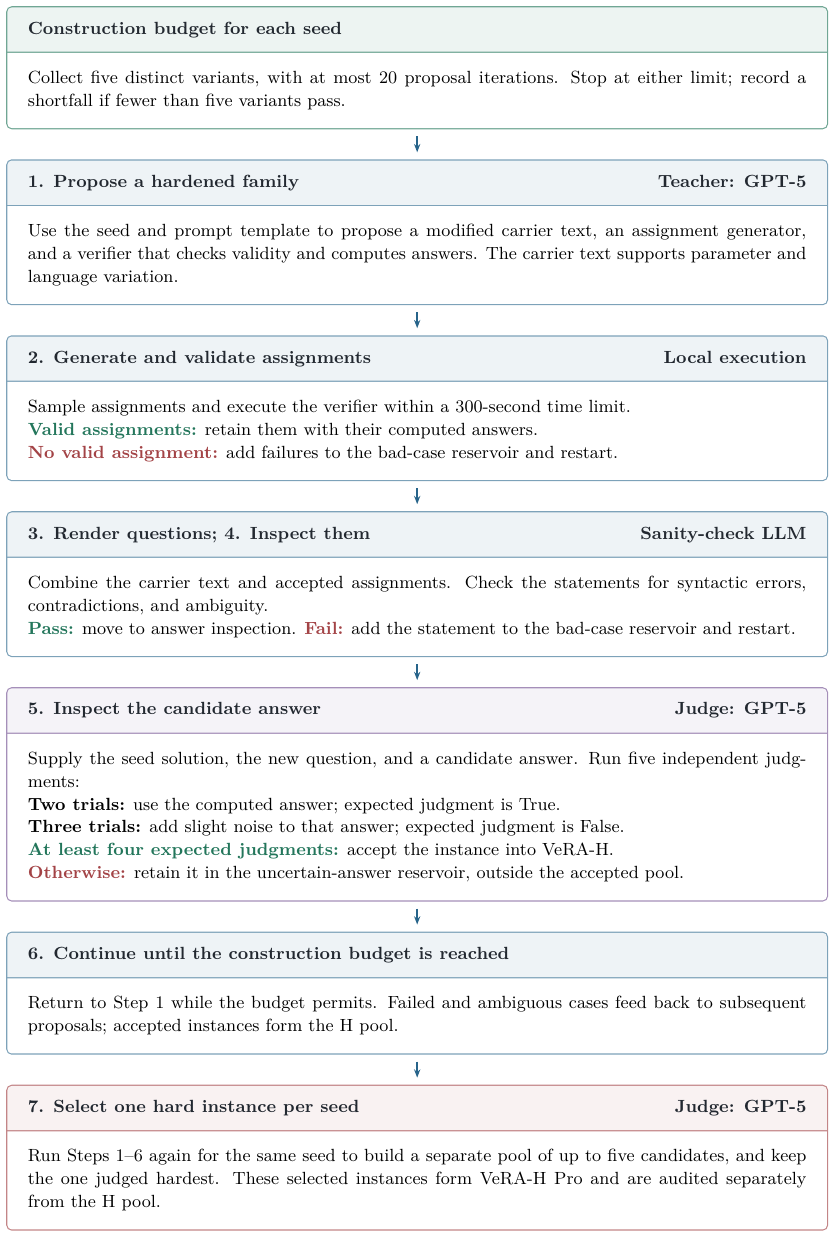}
\caption{\textbf{The H and H Pro construction loop in detail.} Noise-discrimination trials test the proposed labels, and for H Pro a difficulty judge selects one candidate from a separately generated pool. Items that pass these gates then go through the expert audit and repair reported in Table~\ref{tab:hardening}. In the figure, ``variants'' are H candidates, ``carrier text'' is the template and ``verifier'' the answer program.}
\label{fig:workflow-h}
\end{figure}

This appendix illustrates the pipeline on one concrete seed. Figures~\ref{fig:mode-comparison} and~\ref{fig:synthesis-pipeline} follow one geometry seed, AIME 2024 II Problem 10, through \ve{}, \vh{} and \vp{} selection, and Figures~\ref{fig:workflow-e} and~\ref{fig:workflow-h} give the E and H construction loops step by step. In the figures, ``verifier'' denotes the answer program $V_i$ and ``carrier text'' the template $T_i$.

\FloatBarrier
\section{Noise Discrimination and Expert Audit}
\label{app:audit}
This appendix details the two checks on labels that execution cannot provide: noise discrimination for hardened candidates and the expert audit of rendered items.

\subsection{Noise Discrimination}
\label{app:noise}
Noise discrimination tests whether an answer judge can tell the program's answer apart from perturbed alternatives. The judge sees five trials in shuffled order, two with the program's answer and three with perturbed answers, and the candidate passes only if at least four are classified correctly. Integer answers are perturbed by nonzero offsets and real-valued answers by changes proportional to their scale, so the check tests whether the judge is sensitive to answer errors, not merely whether it agrees with a proposed answer.

We calibrated the threshold on 50 AIME items with human judgments. Requiring four correct classifications produced one false acceptance and three false rejections; requiring all five produced zero false acceptances but fourteen false rejections. Four of five therefore keeps most valid candidates, and the expert audit catches the rare false acceptance.

\subsection{Expert Audit}
Section~\ref{sec:validation} describes the audit procedure. Each audited unit is a rendered question together with its answer, and each accepted item is recorded with its specification and audit outcome.

For fresh instances, the audit covers 60 AIME 2024 variants, 60 AIME 2025 variants and 100 Beyond-AIME items, 11 of which are fallback restatements, and all 220 pass. The bounds in Section~\ref{sec:construction} are exact one-sided 95\% Clopper--Pearson upper bounds for zero observed errors. The 1.4\% bound treats the 209 generated variants as independent draws. Variants of one family share an answer program, so their errors may be correlated; counting each of the 149 families once (30 for each AIME year and 89 for Beyond-AIME) gives 2.0\%.

For hardening, the audit covers all seven source--mode groups in Table~\ref{tab:hardening}, including historical AIME: 1,733 of 2,299 candidates pass the initial audit and 454 more pass after repair, for 2,187 accepted items. Section~\ref{sec:construction} describes the three kinds of error the audit catches and how repair fixes each at its source.

\FloatBarrier

\FloatBarrier
\section{Scores, Intervals and Fits}
\label{app:estimands}
This appendix defines the quantities behind the main-text comparisons. Model scores are the Avg@5 accuracies of Equation~\eqref{eq:avg5}, in which every item receives equal weight, so a family contributes in proportion to its number of evaluated items; the panel mean averages these scores over the 16 models. Hardening levels compare the seeds with the Verified sets (sizes in Table~\ref{tab:hardening}), except for historical AIME, which was evaluated only on its raw H pool. Selection gaps (H Pro minus H) are reported on both the raw pools and the Verified sets, with paired intervals on the raw pools.

\paragraph{Intervals.}
Every interval in the main text is a 95\% t-interval with 15 degrees of freedom over the 16 per-model differences, such as a model's \ve{} score minus its seed score. Models are paired units, so inference concerns this panel. The 13.2-point contrast between AIME 2024 and Beyond-AIME in Section~\ref{sec:fresh} applies the same interval to each model's difference between its two changes.

\paragraph{Certifiable difficulty level.}
The line in Section~\ref{sec:hardening-results} is a least-squares fit of panel-mean H accuracy on panel-mean seed accuracy over the four hardening sources, using the Verified H sets and the raw H pool of historical AIME. It gives H $\approx 30+0.47\times$ seed and crosses the diagonal, where the two are equal, at about 57\%. The same fit of Verified H Pro accuracy over its three sources crosses at 47.5\%.

\FloatBarrier
\section{Fresh Instances: Model-Level Results and Edit Size}
\label{app:models}
This appendix gives each model's fresh-instance scores and measures how strongly the variants are edited.

\subsection{Model-Level Results}
Tables~\ref{tab:models-gsm-aime24} and~\ref{tab:models-aime25-beyond} give each model's Avg@5 accuracy on the seeds and on their \ve{} variants, behind the panel means in Table~\ref{tab:equivalent} and Figure~\ref{fig:diagnostics}a. Entries are rounded, so differences computed from them can differ by 0.1 point from those computed before rounding.
\begin{table}[htbp]
\centering\footnotesize
\caption{\textbf{Fresh instances on GSM8K and AIME 2024.} Avg@5 accuracy (\%) of each model on the seeds and on their \ve{} variants (2,638 for GSM8K and 60 for AIME 2024). Panel means and intervals are in Table~\ref{tab:equivalent}.}\label{tab:models-gsm-aime24}
\begin{tabular}{lrrrr}
\toprule
Model & \multicolumn{2}{c}{GSM8K} & \multicolumn{2}{c}{AIME 2024} \\ 
 & Seed & Variant & Seed & Variant \\
\midrule
Gemini 2.5 Pro & 93.9 & 95.0 & 88.0 & 73.3 \\
GLM-4.6 & 94.1 & 94.8 & 90.7 & 71.7 \\
GPT-5.1 high & 95.6 & 96.9 & 92.0 & 75.3 \\
Claude Sonnet 4.5 Thinking & 95.3 & 96.2 & 84.0 & 74.0 \\
Seed-1.6-Thinking-0715 & 95.1 & 95.9 & 85.3 & 72.0 \\
DeepSeek V3.2 Thinking & 94.0 & 95.4 & 92.0 & 79.0 \\
Seed-1.6-1015 high & 95.6 & 96.3 & 89.3 & 75.0 \\
Kimi K2 Thinking & 95.0 & 86.6 & 86.7 & 75.3 \\
MiniMax M2 & 94.7 & 94.9 & 84.7 & 55.7 \\
Gemini 3 Pro Preview & 94.5 & 95.0 & 94.0 & 81.3 \\
GPT-5 high & 95.6 & 97.1 & 90.0 & 74.7 \\
Kimi K2-0905 & 95.0 & 95.6 & 65.3 & 51.3 \\
Qwen3-Max-0923 & 95.5 & 96.9 & 82.7 & 68.7 \\
GPT-5.1-chat-latest & 94.3 & 96.0 & 48.7 & 46.0 \\
Seed-1.6-Lite-1015 high & 95.1 & 95.8 & 87.3 & 73.0 \\
DeepSeek V3.1 Thinking & 94.5 & 94.8 & 90.7 & 77.7 \\
\bottomrule
\end{tabular}
\end{table}

\begin{table}[htbp]
\centering\footnotesize
\caption{\textbf{Fresh instances on AIME 2025 and Beyond-AIME.} Avg@5 accuracy (\%) of each model on the seeds and on their \ve{} variants (60 for AIME 2025; 100 for Beyond-AIME, 11 of which restate their seed). Panel means and intervals are in Table~\ref{tab:equivalent}.}\label{tab:models-aime25-beyond}
\begin{tabular}{lrrrr}
\toprule
Model & \multicolumn{2}{c}{AIME 2025} & \multicolumn{2}{c}{Beyond-AIME} \\ 
 & Seed & Variant & Seed & Variant \\
\midrule
Gemini 2.5 Pro & 85.3 & 73.0 & 57.2 & 57.0 \\
GLM-4.6 & 92.0 & 76.3 & 68.2 & 66.4 \\
GPT-5.1 high & 88.7 & 88.3 & 71.6 & 73.0 \\
Claude Sonnet 4.5 Thinking & 78.7 & 68.7 & 51.6 & 53.0 \\
Seed-1.6-Thinking-0715 & 80.0 & 75.3 & 54.4 & 55.8 \\
DeepSeek V3.2 Thinking & 90.0 & 87.7 & 69.4 & 67.6 \\
Seed-1.6-1015 high & 80.7 & 76.7 & 58.8 & 58.2 \\
Kimi K2 Thinking & 78.7 & 71.7 & 54.8 & 44.0 \\
MiniMax M2 & 76.7 & 57.0 & 56.8 & 54.4 \\
Gemini 3 Pro Preview & 90.7 & 85.7 & 76.6 & 74.0 \\
GPT-5 high & 89.3 & 85.3 & 68.0 & 69.8 \\
Kimi K2-0905 & 47.3 & 44.0 & 37.0 & 34.6 \\
Qwen3-Max-0923 & 74.7 & 69.7 & 53.6 & 58.8 \\
GPT-5.1-chat-latest & 50.7 & 41.7 & 28.8 & 30.6 \\
Seed-1.6-Lite-1015 high & 77.3 & 70.0 & 54.8 & 53.6 \\
DeepSeek V3.1 Thinking & 86.7 & 83.0 & 71.8 & 66.0 \\
\bottomrule
\end{tabular}
\end{table}

\subsection{Edit Size Across AIME Years}
\label{app:perturbation}
The perturbation score $\pi$ measures how far each variant departs from its seed. It is the mean of two distances: numerical distance averages the relative slot changes $|\theta_j-\theta_j^{\rm seed}|/\max(|\theta_j^{\rm seed}|,1)$, clipped to $[0,1]$, and textual distance is the token-level Levenshtein distance divided by the length of the seed's rendering.

\begin{figure}[htbp]
\centering
\includegraphics[width=.65\linewidth]{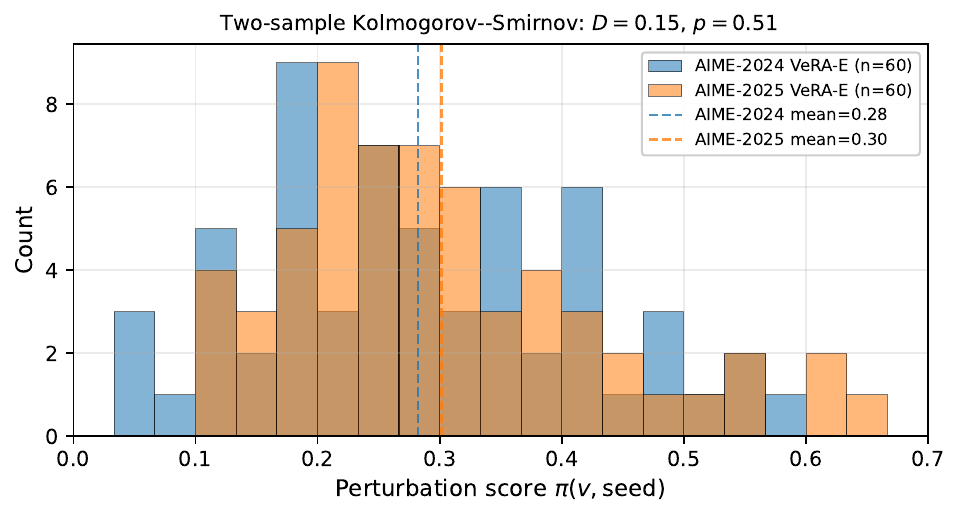}
\caption{\textbf{The two AIME years are edited by similar amounts, yet AIME 2024 drops twice as much.} Distribution of the perturbation score, which combines numerical-slot and textual edit distances, over the 60 variants per year; the means are 0.28 for AIME 2024 and 0.30 for AIME 2025.}
\label{fig:perturbation}
\end{figure}

Figure~\ref{fig:perturbation} shows the distribution of $\pi$ for the two AIME years. The edits are of similar size (variant-level Kolmogorov--Smirnov $D=0.15$, $p=0.51$), yet AIME 2024 drops twice as much as AIME 2025 (Figure~\ref{fig:diagnostics}a). Edit size thus measures how much a variant differs from its seed, not how much harder it is to solve.

\subsection{Worked Failures}
\label{app:worked-failures}
\begin{figure}[H]
\centering
\begin{minipage}[c]{.58\linewidth}
\includegraphics[width=\linewidth]{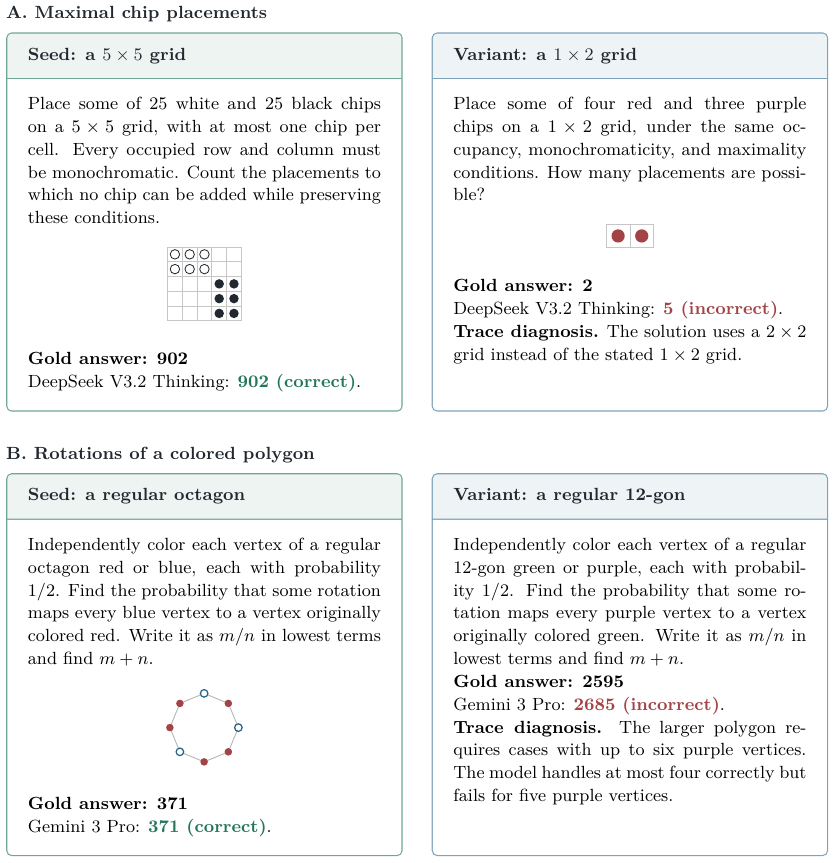}
\end{minipage}\hfill
\begin{minipage}[c]{.39\linewidth}
\caption{\textbf{A model can solve a seed yet fail its variant.} DeepSeek V3.2 Thinking solves the $5\times5$ chip seed but treats the $1\times2$ variant as a square $2\times2$ grid, as if the seed's square shape still applied. Gemini 3 Pro solves the octagon seed, whose case analysis stops at four vertices of one color. The 12-gon variant needs cases up to six, but the model handles only those up to four and fails at five. Both reuse the shape of the seed's solution outside its range.}
\label{fig:paired-examples}
\end{minipage}
\end{figure}

Figure~\ref{fig:paired-examples} gives the full statements of the two failures in Figure~\ref{fig:teaser}b, with a diagnosis from each model's solution trace.

\FloatBarrier
\section{GSM8K Item-Quality Probe}
\label{app:gsm-probe}
This probe tests whether rewriting removes flawed GSM8K items, using the share of items that two strong models both miss as a proxy for flawed statements or labels. GPT-5 and Gemini 2.5 Pro each answered the 1,319 GSM8K test seeds and one sampled variant per seed, with a single response per item (Table~\ref{tab:gsm-probe}).

\begin{table}[htbp]
\centering\small
\caption{\textbf{Fewer GSM8K items are missed by both models after rewriting.} Single-response accuracy (\%) of two models on the 1,319 GSM8K test seeds and on one sampled \ve{} variant per seed, and the share of items that both models miss. The probe was run separately from the main evaluation, so its accuracies are not comparable with the Avg@5 scores in Table~\ref{tab:equivalent} or Appendix~\ref{app:models}.}
\label{tab:gsm-probe}
\begin{tabular*}{.7\linewidth}{@{\extracolsep{\fill}}lrr@{}}
\toprule
 & Seed & Variant\\
\midrule
Gemini 2.5 Pro & 97.12 & 98.56\\
GPT-5 & 96.66 & 98.56\\
Missed by both models & 2.12 & 0.76\\
\bottomrule
\end{tabular*}
\end{table}

The number of items that both models miss falls from 28 to 10, or from 2.12\% to 0.76\% of the set: rewriting raises each model's accuracy and makes the two models agree more often on the correct answer. The probe suggests that fresh instances, which expose familiarity on AIME, also improve item quality on GSM8K.

\FloatBarrier
\section{Hardening: Model-Level Results and Effect of the Audit}
\label{app:extended-models}
This appendix gives each model's scores on the raw hardening pools and shows how the audit changes the comparisons.

\subsection{Model-Level Results}
Tables~\ref{tab:historical-aime}--\ref{tab:raw-models-AMO-Bench} give each model's Avg@5 accuracy on the seeds and raw pools of the four hardening sources. The AIME-2024-II seed scores come from the hardening evaluation of those problems, so each hardening comparison uses seed and candidate scores from one evaluation; Table~\ref{tab:models-gsm-aime24} reports the separate AIME 2024 seed evaluation.

\begin{table}[htbp]
\centering\small
\setlength{\tabcolsep}{4pt}
\caption{\textbf{Historical AIME (1983--2001), raw pool.} Avg@5 accuracy (\%) on the 265 seeds and the 1,321 H candidates before audit; 1,284 candidates are accepted after audit and repair. Historical AIME has no H Pro set.}
\label{tab:historical-aime}
\begin{tabular}{lrr}
\toprule
Model & Seed & H \\
\midrule
Gemini 2.5 Pro & 95.4 & 79.5 \\
GLM-4.6 & 93.8 & 75.6 \\
GPT-5.1 high & 95.2 & 78.4 \\
Claude Sonnet 4.5 Thinking & 94.0 & 75.8 \\
Seed-1.6-Thinking-0715 & 93.8 & 79.1 \\
DeepSeek V3.2 Thinking & 95.7 & 77.7 \\
Seed-1.6-1015 high & 94.3 & 79.1 \\
Kimi K2 Thinking & 59.6 & 71.4 \\
MiniMax M2 & 90.9 & 74.7 \\
Gemini 3 Pro Preview & 93.4 & 74.7 \\
GPT-5 high & 94.5 & 77.4 \\
Kimi K2-0905 & 84.1 & 53.7 \\
Qwen3-Max-0923 & 92.2 & 74.2 \\
GPT-5.1-chat-latest & 72.1 & 55.3 \\
Seed-1.6-Lite-1015 high & 93.1 & 76.3 \\
DeepSeek V3.1 Thinking & 94.4 & 75.3 \\
\bottomrule
\end{tabular}
\end{table}

\begin{table}[htbp]
\centering\small
\setlength{\tabcolsep}{4pt}
\caption{\textbf{AIME-2024-II, raw pools.} Avg@5 accuracy (\%) on the 14 seeds and on the 70 H and 14 H Pro candidates before audit. Panel means for the raw pools and Verified sets are in Table~\ref{tab:raw-verified}. Seed scores come from the hardening evaluation and can differ from the AIME 2024 seed scores in Table~\ref{tab:models-gsm-aime24}.}
\label{tab:raw-models-AIME-2024-II}
\begin{tabular}{lrrr}
\toprule
Model & Seed & H & H Pro \\
\midrule
Gemini 2.5 Pro & 92.9 & 82.0 & 65.7 \\
GLM-4.6 & 94.3 & 81.1 & 67.1 \\
GPT-5.1 high & 94.3 & 90.9 & 74.3 \\
Claude Sonnet 4.5 Thinking & 90.0 & 72.0 & 60.0 \\
Seed-1.6-Thinking-0715 & 87.1 & 76.6 & 67.1 \\
DeepSeek V3.2 Thinking & 91.4 & 84.0 & 68.6 \\
Seed-1.6-1015 high & 92.9 & 76.6 & 67.1 \\
Kimi K2 Thinking & 85.7 & 15.1 & 45.7 \\
MiniMax M2 & 82.9 & 64.3 & 51.4 \\
Gemini 3 Pro Preview & 75.7 & 74.6 & 62.9 \\
GPT-5 high & 88.6 & 82.3 & 78.6 \\
Kimi K2-0905 & 71.4 & 40.6 & 32.9 \\
Qwen3-Max-0923 & 80.0 & 59.1 & 51.4 \\
GPT-5.1-chat-latest & 52.9 & 34.3 & 28.6 \\
Seed-1.6-Lite-1015 high & 88.6 & 70.3 & 57.1 \\
DeepSeek V3.1 Thinking & 90.0 & 81.1 & 58.6 \\
\bottomrule
\end{tabular}
\end{table}

\begin{table}[htbp]
\centering\small
\setlength{\tabcolsep}{4pt}
\caption{\textbf{Beyond-AIME, raw pools.} Avg@5 accuracy (\%) on the 100 seeds and on the 500 H and 100 H Pro candidates before audit. Panel means for the raw pools and Verified sets are in Table~\ref{tab:raw-verified}.}
\label{tab:raw-models-Beyond-AIME}
\begin{tabular}{lrrr}
\toprule
Model & Seed & H & H Pro \\
\midrule
Gemini 2.5 Pro & 57.2 & 57.7 & 52.2 \\
GLM-4.6 & 68.2 & 67.1 & 63.6 \\
GPT-5.1 high & 71.6 & 77.3 & 75.2 \\
Claude Sonnet 4.5 Thinking & 51.6 & 56.3 & 50.2 \\
Seed-1.6-Thinking-0715 & 54.4 & 57.9 & 51.6 \\
DeepSeek V3.2 Thinking & 69.4 & 68.6 & 64.2 \\
Seed-1.6-1015 high & 58.8 & 61.8 & 55.0 \\
Kimi K2 Thinking & 54.8 & 43.8 & 46.6 \\
MiniMax M2 & 56.8 & 52.8 & 47.6 \\
Gemini 3 Pro Preview & 76.6 & 72.2 & 65.4 \\
GPT-5 high & 68.0 & 77.6 & 70.6 \\
Kimi K2-0905 & 37.0 & 30.4 & 25.8 \\
Qwen3-Max-0923 & 53.6 & 60.1 & 52.2 \\
GPT-5.1-chat-latest & 28.8 & 33.0 & 32.6 \\
Seed-1.6-Lite-1015 high & 54.8 & 55.6 & 48.6 \\
DeepSeek V3.1 Thinking & 71.8 & 68.6 & 62.2 \\
\bottomrule
\end{tabular}
\end{table}

\begin{table}[htbp]
\centering\small
\setlength{\tabcolsep}{4pt}
\caption{\textbf{AMO-Bench, raw pools.} Avg@5 accuracy (\%) on the 50 seeds and on the 244 H and 50 H Pro candidates before audit. Panel means for the raw pools and Verified sets are in Table~\ref{tab:raw-verified}.}
\label{tab:raw-models-AMO-Bench}
\begin{tabular}{lrrr}
\toprule
Model & Seed & H & H Pro \\
\midrule
Gemini 2.5 Pro & 28.0 & 43.5 & 37.6 \\
GLM-4.6 & 40.0 & 52.9 & 48.0 \\
GPT-5.1 high & 56.0 & 64.6 & 58.8 \\
Claude Sonnet 4.5 Thinking & 18.0 & 36.0 & 32.4 \\
Seed-1.6-Thinking-0715 & 40.0 & 41.6 & 38.0 \\
DeepSeek V3.2 Thinking & 28.0 & 51.6 & 46.0 \\
Seed-1.6-1015 high & 48.0 & 43.1 & 36.4 \\
Kimi K2 Thinking & 22.0 & 26.5 & 17.2 \\
MiniMax M2 & 20.0 & 36.6 & 28.4 \\
Gemini 3 Pro Preview & 56.0 & 61.1 & 56.0 \\
GPT-5 high & 40.0 & 58.4 & 54.4 \\
Kimi K2-0905 & 8.0 & 19.9 & 18.8 \\
Qwen3-Max-0923 & 14.0 & 39.2 & 35.2 \\
GPT-5.1-chat-latest & 6.0 & 20.7 & 17.6 \\
Seed-1.6-Lite-1015 high & 36.0 & 39.7 & 36.4 \\
DeepSeek V3.1 Thinking & 48.0 & 52.1 & 52.8 \\
\bottomrule
\end{tabular}
\end{table}

\FloatBarrier
\subsection{Effect of the Audit}
\label{app:automatic}
Table~\ref{tab:raw-verified} reports panel means for both the raw pools and the Verified sets. Auditing shifts score levels slightly but leaves every comparison unchanged: AIME-2024-II gains headroom over its seeds, Beyond-AIME gains headroom mainly through H Pro selection, and on AMO-Bench H Pro is harder than H but easier than the seeds. On AIME-2024-II, the raw and Verified H Pro means coincide at 58.57\% because repair changed only three H Pro items, which no model solved before or after repair.

The audit also confirms that the selection gap reflects difficulty. H Pro candidates pass the initial audit less often than H candidates on every source (78.6\% versus 90.0\% on AIME-2024-II, 69.0\% versus 73.0\% on Beyond-AIME and 54.0\% versus 63.5\% on AMO-Bench). Because a flawed label lowers every model's score, part of the raw gap could reflect flaws rather than difficulty. Yet the gap persists on the Verified sets, which contain only accepted items: there H Pro remains harder than H by 10.1, 4.4 and 4.4 points, against 9.2, 4.8 and 4.6 on the raw pools.

\FloatBarrier
\section{Robustness to Excluding Kimi K2 Thinking}
\label{app:robustness}
This appendix checks that no conclusion depends on a single model. Kimi K2 Thinking scores erratically across sources: 15.1\% on the raw AIME-2024-II H pool but 45.7\% on the harder raw H Pro items, 59.6\% on historical AIME seeds where the other 15 models average 91.8\%, and the only GSM8K drop under \ve{}. Table~\ref{tab:robustness} repeats the main comparisons without it. Every conclusion holds, and two sharpen: the AIME-2024-II selection gap widens from 9.2 to 11.9 points with a narrower interval, and GSM8K then improves for every model, as an easy control should.

\begin{table}[htbp]
\centering\small
\caption{\textbf{Main comparisons with and without Kimi K2 Thinking.} Each cell gives the mean change in points, rounded to one decimal, with its 95\% interval over models and the number of models whose accuracy falls; raw denotes the raw pools before audit.}
\label{tab:robustness}
\begin{tabular*}{\linewidth}{@{\extracolsep{\fill}}lll@{}}
\toprule
Comparison & 16 models & 15 models\\
\midrule
GSM8K, E $-$ seed & +0.3 [$-$0.9, +1.6], 1/16 & +0.9 [+0.7, +1.2], 0/15\\
Beyond-AIME, E $-$ seed & $-$1.0 [$-$3.0, +0.9], 10/16 & $-$0.4 [$-$1.8, +1.1], 9/15\\
AIME 2025, E $-$ seed & $-$7.1 [$-$9.8, $-$4.3], 16/16 & $-$7.1 [$-$10.1, $-$4.1], 15/15\\
AIME 2024, E $-$ seed & $-$14.2 [$-$17.0, $-$11.4], 16/16 & $-$14.4 [$-$17.4, $-$11.4], 15/15\\
AIME-2024-II, H Pro $-$ H (raw) & $-$9.2 [$-$15.4, $-$3.1], 15/16 & $-$11.9 [$-$14.6, $-$9.2], 15/15\\
Beyond-AIME, H Pro $-$ H (raw) & $-$4.8 [$-$6.3, $-$3.3], 15/16 & $-$5.3 [$-$6.5, $-$4.2], 15/15\\
AMO-Bench, H Pro $-$ H (raw) & $-$4.6 [$-$5.9, $-$3.3], 15/16 & $-$4.3 [$-$5.5, $-$3.1], 14/15\\
\bottomrule
\end{tabular*}
\end{table}

\FloatBarrier
\section{GPQA-Diamond Extension}
\label{app:gpqa}
GPQA-Diamond \citep{rein2024gpqa} tests the pipeline beyond computable answers. Its multiple-choice answers are not computed by a program, so language models validate them instead. The 198 seeds produce 871 variants, and the 16-model mean is 79.27\% on the seeds and 79.42\% on the variants (Table~\ref{tab:gpqa-models}). Accuracy is essentially unchanged. Because its labels do not come from an answer program, and its variants keep the multiple-choice format, we report this extension apart from the executable mathematical families and do not use it as a familiarity control.

\begin{table}[htbp]
\centering\small
\setlength{\tabcolsep}{4pt}
\caption{\textbf{GPQA-Diamond.} Avg@5 accuracy (\%) of each model on the 198 seeds and their 871 variants, whose answers are validated by language models rather than an answer program (Appendix~\ref{app:gpqa}).}
\label{tab:gpqa-models}
\begin{tabular}{lrr}
\toprule
Model & Seed & Variant \\
\midrule
Gemini 2.5 Pro & 82.1 & 81.4 \\
GLM-4.6 & 78.8 & 80.2 \\
GPT-5.1 high & 86.4 & 84.5 \\
Claude Sonnet 4.5 Thinking & 81.0 & 80.1 \\
Seed-1.6-Thinking-0715 & 78.4 & 78.5 \\
DeepSeek V3.2 Thinking & 82.1 & 83.3 \\
Seed-1.6-1015 high & 77.4 & 78.9 \\
Kimi K2 Thinking & 79.9 & 83.4 \\
MiniMax M2 & 76.7 & 74.4 \\
Gemini 3 Pro Preview & 86.7 & 85.7 \\
GPT-5 high & 81.9 & 82.1 \\
Kimi K2-0905 & 73.5 & 74.2 \\
Qwen3-Max-0923 & 76.1 & 76.7 \\
GPT-5.1-chat-latest & 69.9 & 69.1 \\
Seed-1.6-Lite-1015 high & 77.6 & 76.9 \\
DeepSeek V3.1 Thinking & 80.0 & 81.3 \\
\bottomrule
\end{tabular}
\end{table}

\FloatBarrier
\section{Answer Verification by Structured Debate}
\label{app:debate}
This side study tests whether structured debate among models classifies candidate answers more reliably than a single judge, which would make it a natural upgrade for the answer judge in noise discrimination. It is run independently of noise discrimination and of the 16-model evaluation.

\subsection{Evaluation Set and Decision Target}
The study uses 195 candidate answers to 58 hard competition problems in combinatorics, number theory and algebra. Each problem has one hand-checked correct candidate and two or three incorrect ones, giving 58 positives and 137 negatives. A problem counts as completely classified only when its correct candidate is accepted and all its incorrect candidates are rejected. Both kinds of error matter for construction: a false acceptance puts a wrong label into the accepted pool, and a false rejection shrinks the pool, so precision and recall measure different requirements.

\subsection{Agents, Exchanges and Acceptance}
Both the single-judge baseline and the debate system use GPT-5.2 at low reasoning effort. The debate system assigns five roles: algebraic and formal consistency, theorem applicability, independent resolution, boundary and optimality checking, and numerical or symbolic cross-checking. The five agents share one model but receive different verification instructions.

Each agent first restates the claim, lists possible failure modes, performs an independent check, considers the strongest opposing argument, and records a structured assessment. The assessment distinguishes three outcomes: the answer is supported by evidence, the answer is refuted by evidence, or the reasoning is insufficient but not refuted. This keeps uncertainty alone from counting as proof that an answer is wrong.

The debate runs for up to five rounds. In each round, agents see the vote tally, a digest of unresolved disagreements, challenges generated by a scheduler and messages addressed to them by other agents, and each agent must answer an opposing claim and perform a new concrete check. Previous assessments are carried forward as summaries of at most 400 characters per rationale. A unanimous vote can end the debate after at least two full rounds, and an answer is accepted only if at least three agents explicitly assess it as supported.

\subsection{Results}
\begin{table}[htbp]
\centering\small
\setlength{\tabcolsep}{4pt}
\caption{\textbf{Debate raises precision over a single judge.} Classification of 195 candidate answers to 58 problems in three runs of each system; the baseline is one GPT-5.2 judge. Positive means a correct candidate answer: FP counts incorrect answers accepted and FN correct answers rejected. Accuracy, precision and recall (\%) are computed from these counts. A problem is complete when its correct candidate is accepted and every incorrect candidate is rejected.}
\label{tab:verification-detail}
\begin{tabular*}{\linewidth}{@{\extracolsep{\fill}}lrrrrrr@{}}
\toprule
System/run & Acc. & Prec. & Recall & FP & FN & Complete \\
\midrule
Baseline 1 & 72.8 & 54.5 & 51.7 & 25 & 28 & 18/58 \\
Baseline 2 & 72.8 & 54.9 & 48.3 & 23 & 30 & 18/58 \\
Baseline 3 & 73.8 & 55.9 & 56.9 & 26 & 25 & 22/58 \\
Debate 1 & 87.2 & 90.2 & 63.8 & 4 & 21 & 34/58 \\
Debate 2 & 87.2 & 92.3 & 62.1 & 3 & 22 & 34/58 \\
Debate 3 & 88.7 & 95.0 & 65.5 & 2 & 20 & 36/58 \\
\bottomrule
\end{tabular*}
\end{table}

Across three independent runs, mean precision rises from 55.1\% for the single judge to 92.5\% for debate, mean false acceptances fall from 24.7 to 3.0 per run, and mean false rejections fall from 27.7 to 21.0 (Table~\ref{tab:verification-detail}). Debate classifies 34.7 of the 58 problems completely, compared with 19.3 for the single judge, so it accepts a much cleaner answer pool while also accepting more correct answers.

The remaining false rejections fall into three groups: the hardest solutions were difficult to check, agents raised incorrect mathematical objections, or the evidence was mixed and fell short of the acceptance threshold. A good judge must therefore be able to confirm a difficult but valid argument, not only detect a flawed one. Debate uses up to 25 calls per candidate, against one for the single judge, so the gain reflects the full system: structured exchange together with a larger verification budget.

\FloatBarrier
\section{Training Use and Held-Out Evaluation}
\label{app:exposure}
Because code labels every draw, a specification can also generate labeled training data, and evaluation then needs a held-out split. Each specification records the seed identity, slot schema, template, generator, answer program, version, sampling index and validation outcomes, so every question and its answer can be regenerated and every split documented. Evaluation data can be held out in two ways. Holding out sampling indices gives new instances of computations the model has seen, whereas holding out entire families also changes the computation being tested, so the two splits answer different generalization questions. In either case, the evaluation should document the exposure and keep the held-out families or input regimes separate.

\FloatBarrier
\section{Extended Related Work}
\label{app:research-context}
This appendix extends Section~\ref{sec:related} theme by theme.

\paragraph{Refreshing instances and extending difficulty.}
Contamination studies examine the relationship between benchmark exposure and measured performance \citep{shi2023detecting,golchin2024time,sainz2023contamination}, and LiveCodeBench and LiveBench renew evaluation material over time \citep{jain2024livecodebench,white2024livebench}. Perturbation-based evaluations ask a complementary question: how well does a solver handle changes to a known problem family? GSM-Plus varies mathematical statements, GSM-Symbolic and VAR-MATH parameterize instances, and GSM-Ranges varies numerical scale \citep{li2024gsmplus,mirzadeh2024gsm,yao2025varmath,shrestha2025gsmranges}. \ve{} expresses the relationship between instances through a reusable program, \vh{} also changes the computation, and \vp{} selects among the resulting candidates, so one construction interface delivers both freshness and headroom.

\paragraph{Exposure diagnostics.}
Surveys of benchmark contamination distinguish detection approaches and sources of evaluation leakage \citep{xu2024contaminationsurvey}, and tests such as PaCoST and GSM1k compare performance on familiar and newly constructed material \citep{zhang2024pacost,zhang2024gsm1k}. These studies motivate new tests whose relationship to the source can be inspected. \vera{} makes that relationship explicit: the seed identifies the originating problem, the generator defines valid changes, and the answer program labels each resulting instance, so repeated sampling keeps a concrete link to the computation being evaluated.

\paragraph{Functional benchmarks and procedural generation.}
Procgen evaluates reinforcement-learning generalization with procedurally generated environments \citep{cobbe2020procgen}. In mathematics, Putnam-AXIOM combines static and functional evaluation, and RV-Bench introduces questions with unseen random-variable assignments \citep{gulati2025putnam,hong2025rvbench}. Functional MATH hand-writes problem, input-generator and solution functions for part of MATH \citep{srivastava2024functional}, MATH-Perturb pairs problems with annotator-written simple and hard perturbations \citep{huang2025mathperturb}, and TemplateGSM generates grade-school templates automatically \citep{zhang2025templategsm}. BeyondBench generates algorithmic tasks with deterministic solutions \citep{srivastava2025beyondbench}. In all of these, one evaluation family can supply many items instead of one fixed item. \vera{} adds a pipeline that starts from natural-language competition problems, compiles reusable generators and answer programs automatically, and searches for harder computations, so one representation supports both renewal within a family and harder versions of it.

\paragraph{Graphs, rewrites and executable specifications.}
DyVal and DARG manipulate reasoning graphs to vary task structure and complexity \citep{zhu2024dyval,zhang2024darg}. MaSTer uses iterative rewriting and verification to search for robustness failures \citep{hou2025archecker}, and AutoLogi constructs logic puzzles with executable verification \citep{liu2025autologi}. \vera{} represents a family by a natural-language template, a constrained input generator and an answer program. The answer program can use arithmetic, conditionals, iteration and search, so graph composition can be implemented inside it, while the generator determines which assignments are valid and every rendered question must stay consistent with the slots and the computation. One inspectable family thus combines instance generation, answer computation and consistency checks.

\paragraph{Execution-based evaluation of generated programs.}
Codex and program-synthesis studies evaluate programs through their outputs on tests \citep{chen2021codex,austin2021program}, APPS and TACO extend this setting to diverse algorithmic problems \citep{hendrycks2021apps,li2023taco}, and EvalPlus strengthens code evaluation with additional generated tests \citep{liu2023evalplus}. This line of work shows the value of checking behavior across many inputs rather than a single successful example. \vera{} uses execution on the construction side instead: each accepted program becomes a reusable source of labels, whose validity depends on the implementation, the allowed input domain and the agreement between the rendered question and the computation.

\paragraph{Formal verification and executable answers.}
MiniF2F supplies mathematical statements for machine-checked proving, and LeanDojo connects language models to a proof environment and premise retrieval \citep{zheng2022minif2f,yang2023leandojo}. These systems verify derivations against a formal statement, whereas \vera{} targets question--answer instances: the specification computes an answer, and validation checks the generated program and its agreement with the rendered question. Both approaches provide machine-checkable feedback by exposing an explicit object against which a proposed result can be checked.

\paragraph{Model judgment.}
Verifiers, self-consistency and multi-agent debate offer ways to evaluate candidate reasoning without executable labels \citep{cobbe2021training,lightman2023verify,wang2022selfconsistency,du2023debate}. In \vera{}, such judgment complements execution rather than replacing it: noise discrimination screens hardened candidates, the expert audit establishes the validity of each accepted item, and the debate study in Appendix~\ref{app:debate} measures how well candidate answers can be classified.

\paragraph{Generation, selection and designed distributions.}
AlphaCode combines large-scale program generation with filtering and selection, and CodeRL uses execution feedback in learning and candidate refinement \citep{li2022alphacode,le2022coderl}. STaR learns from successful generated reasoning \citep{zelikman2022star}, and DeepSeek-R1 and Absolute Zero use verifiable feedback to improve reasoning, with Absolute Zero also proposing its own code-based tasks \citep{deepseek2025r1,zhao2025absolutezero}. Synthetic training data follows the same pattern: Self-Instruct and WizardLM generate instruction data through proposals, filtering and progressive rewriting \citep{wang2023selfinstruct,xu2023wizardlm}, Orca emphasizes explanation traces, Textbooks Are All You Need studies curated and synthetic educational code data \citep{mukherjee2023orca,gunasekar2023textbooks}, and MetaMath and MAmmoTH bootstrap mathematical questions and combine verbal and program-based rationales \citep{yu2024metamath,yue2024mammoth}. In all of these, what is proposed and how it is selected together determine the useful outputs. \vera{} applies the same principle to evaluation: validation decides which H candidates become usable items, and difficulty-based selection among a seed's candidates produces H Pro, so H and H Pro are two budgeted searches by the same process, one unselected and one selected, and the selection policy shapes how hard the resulting benchmark is.

\paragraph{Acceptance rules and documentation.}
Datasheets for Datasets argues for documenting construction decisions alongside the data \citep{gebru2021datasheets}. In \vera{}, the acceptance rule fixes what counts as a usable item, and seed identifiers, sampling indices, validation outcomes and repair decisions connect each accepted question to its origin. This keeps three steps distinct: proposing many candidates, obtaining valid items, and assembling a benchmark with a particular difficulty profile. The executable family is the reusable unit, and the construction budget determines the particular sets evaluated in this paper.

\end{document}